\documentclass{ecai}  

\usepackage{graphicx}
\usepackage{latexsym}

\usepackage{amssymb}
\usepackage{amsmath}
\usepackage{amsthm}
\usepackage{booktabs}
\usepackage{enumitem}
\usepackage{color}
\usepackage{overpic}
\usepackage{multirow}
\usepackage{colortbl}
\usepackage{bbding}
\usepackage{pifont}

\definecolor{lightgrey}{RGB}{244,244,244}
\definecolor{grey}{RGB}{128,128,128}
\definecolor{midgrey}{RGB}{225,225,225}
\definecolor{forestgreen}{RGB}{47,159,87}
\definecolor{customred}{RGB}{255,0,0}

\newcommand{\cmark}{\textcolor{forestgreen}{\ding{51}}} 
\newcommand{\xmark}{\textcolor{customred}{\ding{55}}}    

\begin{document}

\begin{frontmatter}

\title{OmniCLIP: Adapting CLIP for Video Recognition with \\Spatial-Temporal Omni-Scale Feature Learning}

\author[A]{\fnms{Mushui}~\snm{Liu}}
\author[A]{\fnms{Bozheng}~\snm{Li}}
\author[A]{\fnms{Yunlong}~\snm{Yu}\thanks{Corresponding Author. Email: yuyunlong@zju.edu.cn}} 
\address[A]{Zhejiang University}

\begin{abstract}
Recent Vision-Language Models (VLMs) \textit{e.g.} CLIP have made great progress in video recognition. Despite the improvement brought by the strong visual backbone in extracting spatial features, CLIP still falls short in capturing and integrating spatial-temporal features which is essential for video recognition. In this paper, we propose OmniCLIP, a framework that adapts CLIP for video recognition by focusing on learning comprehensive features encompassing spatial, temporal, and dynamic spatial-temporal scales, which we refer to as omni-scale features. This is achieved through the design of spatial-temporal blocks that include parallel temporal adapters (PTA), enabling efficient temporal modeling. Additionally, we introduce a self-prompt generator (SPG) module to capture dynamic object spatial features. The synergy between PTA and SPG allows OmniCLIP to discern varying spatial information across frames and assess object scales over time. We have conducted extensive experiments in supervised video recognition, few-shot video recognition, and zero-shot recognition tasks. The results demonstrate the effectiveness of our method, especially with OmniCLIP achieving a top-1 accuracy of 74.30\% on HMDB51 in a 16-shot setting, surpassing the recent MotionPrompt approach even with full training data. The code is available at \url{https://github.com/XiaoBuL/OmniCLIP}.
\end{abstract}

\end{frontmatter}


\section{Introduction} \label{sec:intro}

With the surge of large-scale Internet video data, video recognition \cite{vivit,TimeSformer,k400} has become increasingly critical. Recently, image-text pre-training models like CLIP \cite{clip} and ALIGN \cite{ALIGN} have shown remarkable capabilities in the downstream image tasks \cite{rao2022denseclip,coop}, thanks to their robust spatial feature extraction and open-vocabulary functionalities. Nevertheless, developing a similar video model demands significant computational resources. Therefore, there's a growing trend towards adapting pre-trained image-text models like CLIP for video recognition \cite{x-clip,st-adapter,wasim2023vita}. However, the inherent design of CLIP for static images mismatches with the dynamic nature of video, posing two primary challenges in its application for video recognition.

\begin{figure}[!tb]
    \centering
    \begin{overpic}[width=\linewidth]{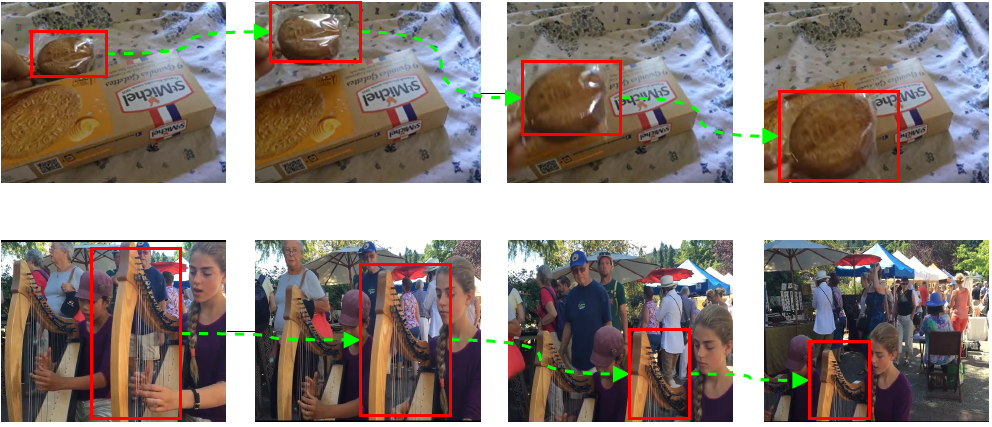} \footnotesize
       \put(10,21){(a) Class: ``Pulling something from behind of something”}
       \put(30,-4){(b) Class: ``playing harp”} 
    \end{overpic}
    \vspace{-3mm}
    \caption{Video recognition is challenging due to the dynamic motion and variations of objects in multi-frames.}
    \label{fig:intro-figure}
\end{figure}

\begin{figure*}[!tb]
    \centering
    \begin{tabular}{cc}
        \begin{overpic}[width=0.70\linewidth, height=0.2\linewidth]{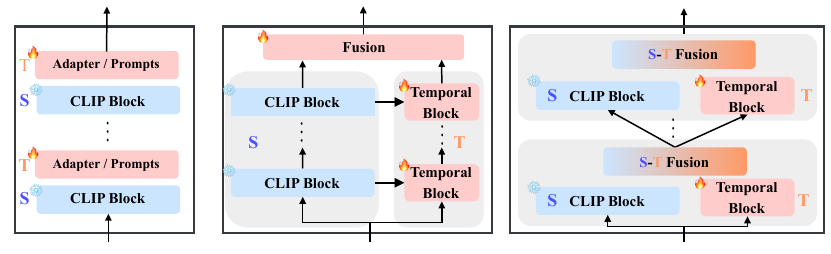} \end{overpic} &
        \begin{overpic}[width=0.25\linewidth, height=0.185\linewidth]{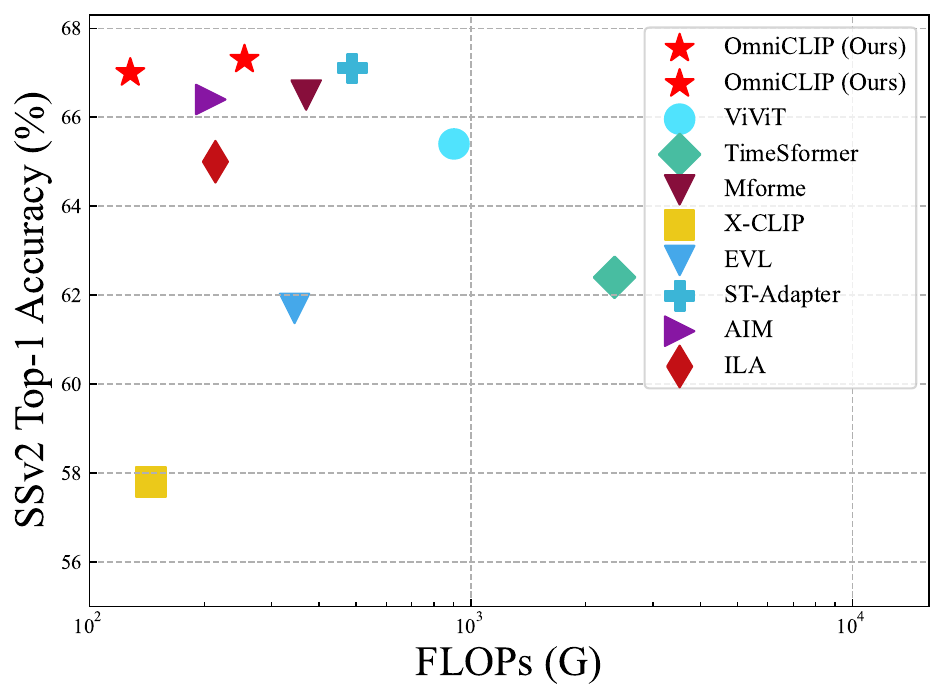} \end{overpic} \\
    \end{tabular}
    \put(-480, -55){\small (a) ST-Adapter} 
    \put(-360, -55){\small (b) EVL}
    \put(-250, -55){\small (c) OmniCLIP (Ours)}
    \put(-120, -55){\small (d) Performance Comparison}
    \caption{Comparison with the recent fine-tuning CLIP to video recognition. S and T refer to spatial modeling and temporal modeling, respectively. (a) ST-Adapter. (b) EVL. (c) OmniCLIP~(Ours) dynamically fuses the spatial-temporal information in parallel. (d) Performance comparison on the SSv2 dataset.}
    \vspace{-2mm}
    \label{fig:intro-comparison}
\end{figure*}

The first challenge is dynamic object tracking, as shown in Fig.~\ref{fig:intro-figure}~(a). This requires the models not only to identify objects within each frame of the video but also to understand the actions across a series of frames over time. However, CLIP, designed for static image-text pairs, struggles with video recognition, as it poorly tracks object motion and frame continuity. The second challenge involves managing the video's continuous nature. Videos, unlike still images, evolve in time, altering object and scene characteristics. The models need to take this into account when recognizing objects, accounting for changes in their size, appearance, and behavior throughout the video, as depicted in Fig.~\ref{fig:intro-figure}~(b). Thus, to address the above challenges arising from the video's dynamic nature, it is crucial to integrate temporal information across frames into CLIP to improve its capability of capturing motion trajectories and frame connections. 

The existing approaches for establishing temporal cues in CLIP involve integrating learnable adapters or prompts, mostly falling into two categories: temporal-embedded and temporal-side models. Temporal-embedded models, \textit{e.g.}, ST-Adapter \cite{st-adapter} and VitaCLIP \cite{wasim2023vita}, which couple space and time, involve sequential temporal-spatial integration adapters or prompts, as depicted in Fig.~\ref{fig:intro-comparison}~(a).  While effective in handling motion and object details, they are at the cost of stacking heavy temporal-spatial modules to capture video information, constrained by lower computational efficiency. On the other hand, temporal-side models like EVL independently develop temporal modeling while maintaining static spatial representations, as demonstrated in Fig.~\ref{fig:intro-comparison}~(b). They offer computational efficiency but may overlook the influence of temporal dynamics on spatial details, potentially leading to suboptimal results. However, both temporal modeling manners face challenges in transferring to dynamic spatial-temporal scales, particularly in managing complex, non-linear spatial-temporal patterns, including irregular movements or the evolution of spatial structures over time.

In this paper, we argue that the integration of temporal information should be \textit{omni-scale}, encompassing a seamless blend of spatial, temporal, and spatial-temporal integration. To achieve this, we present OmniCLIP, a novel approach that adapts CLIP for video recognition through spatial-temporal omni-scale feature learning, enabling a more sophisticated and dynamic understanding of video content, as shown in Fig.~\ref{fig:intro-comparison}~(c). Specifically, OmniCLIP integrates a parallel temporal adapter (PTA), tailored to enhance CLIP's capabilities in temporal modeling. PTA integrates a temporal attention mechanism, meticulously tracking information across frames at consistent spatial locations. Additionally, PTA works parallel with the frozen spatial CLIP block, integrating spatial information through a straightforward learnable addition operation, thereby enabling OmniCLIP to effectively balance temporal adaptation while maintaining computational efficiency. Furthermore, OmniCLIP employs a Self-Prompt Generator (SPG) to effectively handle the dynamic interactions and irregular movements of objects across various spatial scales. Specifically, SPG leverages average pooling and a learnable projector to extract refined multi-scale information and capitalize on CLIP's strong spatial capabilities. Lastly, the combination of the PTA and SPG allows OmniCLIP to distinguish varying spatial information across frames and object scales over time, establishing a robust dynamic and spatial-temporal omni-scale framework for video recognition. This results in an efficient and competitive performance, as demonstrated in Fig.~\ref{fig:intro-comparison}~(d). To summarize, our highlights are as follows:

\begin{itemize}
    \item We propose OmniCLIP, a framework to adapt CLIP for video recognition. OmniCLIP learns omni-scale video representations in spatial, temporal, and spatial-temporal scales.
    \item Our design incorporates a parallel temporal adapter and self-prompt generator to facilitate temporal modeling and dynamic adaptation to spatial-temporal scales. These modules enable OmniCLIP to achieve enhanced efficiency and performance in spatial-temporal processing.
   \item We evaluate proposed OmniCLIP with various settings \textit{i.e.} supervised setting, few-shot setting, and zero-shot setting on different datasets \textit{i.e.} Kinetics-400 \cite{k400}, Something-to-Something v2 \cite{goyal2017something}, HMDB51 \cite{kuehne2011hmdb}, and UCF101 \cite{soomro2012ucf101}. Experimental results have shown the superiority of our method, especially on the few-shot video recognition benchmark.
\end{itemize}

\section{Related Works}
\textbf{Video Recognition.} Video recognition stands as a crucial task within the video domain. The evolution of video recognition techniques has seen significant progress, transitioning from hand-crafted feature-based methods \cite{klaser2008spatio,laptev2005space,wang2013dense}, CNN methods \cite{he2016deep,carreira2017quo,christoph2016spatiotemporal} to existing well-performance Transformer-based methods \cite{vivit,TimeSformer,liu2022videoswin}. However, training video models from scratch is costly and time-consuming due to the large volume of video data. The rising of pretraining visual model \cite{vit,clip} draws great attention under such a situation and gradually grows into the priority choice of the image backbone \cite{aim,st-adapter} for video recognition tasks.

\noindent
\textbf{Vision-Language Model.} Among various pre-training models \cite{jin2024impact,mars}, Vision-Language Models (VLMs) like CLIP \cite{clip} and ALIGN \cite{ALIGN} have shown promising performance in various downstream tasks \cite{coop,ifss,liu2023sync}. CLIP, for example, effectively aligns image and text representations through web-scale image-text pairs in a contrastive manner. Due to the remarkable success of the vision-language pre-training paradigm, some efforts \cite{xu2021videoclip,huang2023clover} have delved into utilizing video-text pairs for pre-training video-language models. OmniVL \cite{wang2022omnivl}, for instance, introduces a unified vision-language contrastive (UniVLC) loss, aiming to maximize the exploitation of information from diverse modalities during model pre-training. While these models exhibit outstanding performance, their pre-training process frequently demands considerable resources. Consequently, methods that leverage image-based VLMs for video applications have gained popularity as they offer an efficient alternative.

\noindent
\textbf{Adapting CLIP for Video Recognition} To tackle the problem of resource limitation, adapting the pre-trained image model \textit{e.g.} CLIP for video tasks can offer an efficient and effective solution. A key aspect of this efficient adaptation process is spatial-temporal modeling. Some works \cite{wu2023revisiting,wu2023bidirectional,wang2021actionclip,x-clip} fully finetune the backbone with a video-specific structure based on CLIP backbone. ActionCLIP \cite{wang2021actionclip} models temporal information at multiple levels including frame and video level, while X-CLIP \cite{x-clip} introduces a cross-frame attention mechanism for temporal information modulation. Open-VCLIP \cite{weng2023open} is designed for tackling open-vocabulary zero-shot tasks \cite{lu2024learning}, involving fine-tuning all parameters of CLIP models. Others \cite{wasim2023vita,aim,st-adapter,evl} utilize PEFT, \textit{e.g.}, adapter structures or learnable prompts, to inject temporal learning ability into CLIP. EVL \cite{evl} utilizes a lightweight transformer decoder to capture temporal interactions among frames, and ST-Adapter \cite{st-adapter} employs an efficient adapter with 3D convolution to concurrently learn spatial-temporal representation. Vita-CLIP \cite{wasim2023vita} introduces three types of prompts to enhance temporal modeling. Prompt-based and adapter-based methods can be more resource-efficient when adapting the CLIP model for video understanding. In this work, we also transfer CLIP model with a PEFT manner to video recognition and capture the omni-scale features by a well-designed temporal adapter and dynamic spatial-temporal feature refinement. Different from TimeSformer \cite{bertasius2021space} and X-CLIP \cite{x-clip}, which also investigate temporal attention, our research focuses on the seamless integration of spatial information from the original CLIP block with temporal information derived from the temporal attention mechanism. This integration is achieved in a parallel fashion, setting our work apart from existing literature in a significant manner.

\section{Method}
\begin{figure*}
    \centering
     \begin{overpic}[width=\linewidth]
     {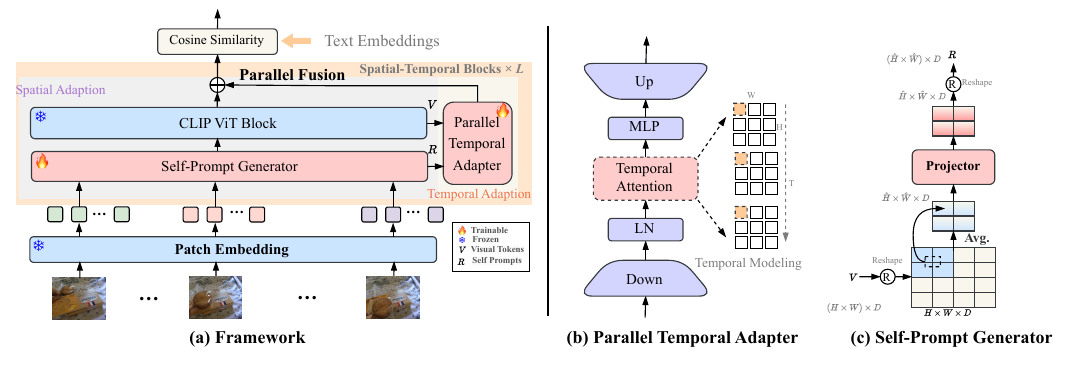}
    \end{overpic}
    \vspace{-8mm}
    \caption{ (a) The framework of our proposed OmniCLIP. (b) PTA establishes the temporal modeling. (c) SPG enhances spatial representations. The combination of PTA and SPG further improves spatial-temporal omni-scale learning.}
    \vspace{-2mm}
    \label{fig:Framework}
\end{figure*}

\subsection{Architecture Overview} \label{sec3.1}
OmniCLIP, designed for dynamic occlusion and temporal variation in video recognition, merges spatial-temporal features for a thorough, multi-scale insight. It primarily comprises a video encoder $\theta_{V}$ and a text encoder $\theta_{T}$. 
The video encoder contains two main components: the parallel temporal adapter (PTA) and the self-prompt generator (SPG). Moreover, leveraging pre-trained image-text alignment, OmniCLIP employs video-specific text features to enhance zero-shot generalization and video-text alignment.

\textbf{Video Encoder} $\theta_{V}$ comprises $L$ spatial-temporal blocks, essential in extracting omni-scale video features, as shown in Fig.~\ref{fig:Framework}. Each block combines a spatial ViT layer, using CLIP pre-training weights fixed during training, with a temporal adapter, actively trained for motion capture. Given a video \( \mathbf{V} \in \mathbb{R}^{T \times H \times W \times 3} \), consisting of \( T \) frames, processes each frame \( t \) (where \( t \in \{1, \ldots, T\} \)) by dividing it into \( K \) non-overlapping square patches of size \( P \times P \) using a ViT architecture \cite{vit}, where the total patch count \( K \) is \( H \times W / P^2 \). Each patch is initially embedded into \( d \)-dimensional features \( X^0 = \{x_{t,j} \in \mathbb{R}^{d} \mid 1 \leq j \leq K,  1 \leq t \leq T \} \), where \( j \) represents the patch number. Next, class tokens \( e_{cls} \) are prepended to the tokens as \( X^0 = \{e_{cls}, X^0\} \). The input for the spatial-temporal blocks, augmented with positional encoding \( \{PE_{i}\}_{i=1}^{N} \) and temporal encoding \( \{TE_{i}\}_{i=1}^{T} \), is formulated as: 

\begin{equation}
\label{eq:pos} 
    V^{0} = X^{0} + PE + TE.
\end{equation} 

Subsequently, a self-prompt generator (SPG) module extracts multi-resolution video information, commencing with self-resolution prompts $R^{0} = {\rm{SPG}}(V^{0})$. The initial video input is subsequently augmented by concatenating it with these prompts, resulting in $V^{0} = {\rm{Concat}}(V^{0}, R^{0})$. Both the spatial ViT layer and the temporal adapter receive the identical input, $V^{0}$. Then, the spatial output of the $i_{th}$ layer can be derived.
\begin{equation}
    \label{eq:spatial}
    V^{i}_s = {\rm{ViT}}(V^{i - 1}).
\end{equation}

In parallel, a learnable temporal adapter is used for extracting the temporal cues as:
\begin{equation}
    \label{eq:spatial}
    V^{i}_t = {\rm{PTA}}(V^{i - 1}).
\end{equation}

Then a simple fusion module fuses temporal and spatial cues as:
\begin{equation}
    \label{eq:fuse}
    V^{i} = V^{i}_s + \alpha * V^{i}_t,
\end{equation}
where $\alpha$ is a learnable factor to balance the two items. Lastly, the class token $e_{cls}^{(N)}$ from the last transformer layer is projected to a latent space with a linear layer to obtain the final frame-level representation $F_{V,t}^{cls}$. Then, the aggregated video-level representation is formulated as:  
\begin{equation}
\label{1} \mathbf{F}_V = {\rm{Avg}}({\rm{MHA}}([F_{V,1}^{cls},F_{V,2}^{cls},...F_{V,t}^{cls},...,F_{V,T}^{cls}])),
\end{equation}
where $\rm{MHA}$ is the multi-head attention layer and $\rm{Avg}$ denotes an average pooling operation. 

Overall, the PTA module is used for temporal modeling and the SPG for spatial refinement. The combination of these two modules enables OmniCLIP to effectively extract omni-scale spatial-temporal features for video recognition.

\textbf{Text Encoder} $\theta_{T}$ consists of several Transformer blocks \cite{vaswani2017attention} and remains frozen throughout the training process. Given a video label $y$ and the class name ``[CLS]", we create a description $T$ using a pre-defined template: ``A video of [CLS] action". Then, we extract the text feature $\mathbf{F}_T$ using the text encoder: $\mathbf{F}_T = \theta_{T} (T)$. Following the approach in \cite{x-clip}, we further enrich the text feature with video-specific prompts.

\subsection{Self-Prompt Generator}\label{sec3.2}
Prompt learning \cite{coop} has been well explored in transferring VLMs. In our work, we introduce a unique prompt designed to enhance the representation of video spatial extraction. Motivation stems from the variation in object resolution under different perspectives within videos. Consequently, capturing information about the resolution of distinct objects is of considerable importance. To address this, we propose a self-prompt generator for augmentation. Specifically, SPG initially employs Average Pooling for the downsampling of video input $V^i$, resulting in down-sampled video features $R^{i} \in \mathbb{R}^{T \times \frac{K}{4} \times d}$. Subsequently, these features undergo a spatial mapping through a projector:
\begin{equation}
    \label{SRPG}
    R^{i} = {\rm{Projector}} ({\rm{Avg}} (V^i))
\end{equation}
where $\rm{Projector}$ consists of two-layer MLPs. Consequently, the self-prompt generator (SPG) can enhance the extraction of spatial features in videos by learning different resolutions with the frozen ViT blocks.

\subsection{Parallel Temporal Adapter} \label{sec3.3}
To adapt the image-based models for video recognition, previous work \cite{st-adapter,wasim2023vita,evl} usually incorporates adapters or prompts to capture the temporal information across frames. However, serial adapter structure \cite{st-adapter} would raise the high computational costs due to the gradient backpropagation while temporal-side fusion mechanism \cite{evl} highly relies on the representation of spatial vision backbone. Our major objectives are to construct an efficient temporal adapter that can extract temporal cues independently and enhance the representation of the origin spatial backbone with \textbf{dual-direction interaction}.

To this end, we propose to build the temporal information via a parallel temporal adapter (PTA). Specifically, PTA, which consists of a tunable self-attention layer to aggregate the same spatial location across \( T \)frames, is wrapped with the bottleneck structure containing the down and up projection. Given the visual tokens $V^{i-1} \in \mathbb{R}^{B\times T\times K\times D}$ in $i^{th}$ layer, where $B$ is the batch size, $T$ is the frames, $K$ is the patch numbers, and $D$ is the feature dimension, PTA firstly shift the visual tokens to $\mathbb{R}^{\left(B K\right) \times \left(T\right) \times D}$, and PTA can extract the temporal knowledge as:
\begin{equation}    
    \label{eq:temporal prompts}
     V_{t}^{i} = {\rm{PTA}}(V^{i-1}) = {\rm{Up}} ( {\rm{Attn}} ( {\rm{Down}}(V^{i-1}))),
\end{equation}
where the Down projector projects the reshaped visual tokens to a low-dimensional space for calculating the motion information and the Up projector restores the refined temporal visual feature. Note that the PTA module consists of the self-attention layers, which share the same structure as the attention block in the ViT \cite{vit} backbone while the weight is randomly initialized and tunable. By connecting the same spatial location in temporal dimensions, PTA can effectively capture the temporal cues in the video.

Note that the input of PTA contains self-prompts $R^i$, as derived by the SPG module described in Eq.~\ref{SRPG} which captures multi-resolution information. Therefore, PTA harnesses the temporal aspects of $R^i$, accumulating large-scale spatial information. This combination of SPG and PTA enables OmniCLIP to access broader spatial windows aross frames and progressively integrate spatial-temporal information.

\subsection{Training Objectives} \label{sec3.4}
Once obtained the video feature representation $\mathbf{F}_V^i$ of video sample $x_i$ and the text feature representation $\mathbf{F}_T^j$ of descriptions of class $c_j$, we calculate their cosine similarity score to quantitatively measure their semantic similarity, i.e., 
\begin{equation}
\label{eq:sim} 
    sim(x_i, c_j) = \frac{\langle \mathbf{F}_V^i, \mathbf{F}_T^j\rangle}{\Vert \mathbf{F}_V^i \Vert \Vert \mathbf{F}_T^j \Vert},  
\end{equation}
where $\langle ~, \rangle$ denotes the inner product operation. The entire model is trained by optimizing the visual-text semantic similarity when the videos and texts belong to the same class, and minimizing it when they belong to different classes. Thus, the objective function is formulated by:

\begin{equation}
\label{eq:loss} 
    \mathcal{L}_{obj} = \sum_{i} -y_i\log {\rm{SM}} ~ sim(x_i, \cdot), 
\end{equation}
where SM denotes the softmax function, $y_i$ denotes the one-hot class label of $x_i$, $sim(x_i, \cdot)$ denotes the semantic similarity vector, where each element represents the semantic similarity score between $x_i$ and a class from the candidate categories.

\definecolor{tabcolor}{RGB}{200, 200, 200}
\def\tabwidth{.31}
\begin{table}[t]
    \centering
    \caption{Implementation details of OmniCLIP under supervised setting.}
    \resizebox{\linewidth}{!}{
    \begin{tabular}{lcccc}
    \toprule
    Implementation Details & K400 & SSv2 & HMDB-51 & UCF-101 \\
    \midrule
    \rowcolor{tabcolor} \textbf{Optimization} & & & & \\
    Batch size & 256 & 256 & 32 & 32 \\ 
    Learning rate & 2e-3 & 3.5e-3 & 2e-3 & 2e-3 \\ 
    Minimal learning rate & 2e-5 & 3.5e-5 & 2e-5 & 2e-5 \\
    Training epochs & 50 & 50 & 30 & 30 \\
    Optimizer & \multicolumn{4}{c}{AdamW} \\
    Learning rate schedule & \multicolumn{4}{c}{Cosine} \\ 
    Learning warmup epochs & \multicolumn{4}{c}{5} \\
    Optimizer betas & \multicolumn{4}{c}{(0.9,0.98)} \\
    \midrule
    \rowcolor{tabcolor} \textbf{Data augmentation} & & & & \\
    RandomFlip & \multicolumn{4}{c}{0.5} \\
    ColorJitter & \multicolumn{4}{c}{0.8} \\
    GrayScale & \multicolumn{4}{c}{0.2} \\
    Label smoothing & \multicolumn{4}{c}{0.1} \\
    Mixup & \multicolumn{4}{c}{0.8} \\ 
    Cutmix & \multicolumn{4}{c}{1.0} \\
    \midrule
    \rowcolor{tabcolor} \textbf{Regularization} & & & & \\
    Weight decay & 0.003 & 0.01 & 0.003 & 0.003 \\
    \bottomrule
    \end{tabular}
    }
    \label{tab:imple}
\end{table}
\section{Experiments}
\definecolor{tabcolor}{RGB}{200, 200, 200}
\def\tabwidth{.31}

In this section, we present an assessment of the efficacy of our approach through its application to three distinct video recognition settings: supervised learning, few-shot learning, and zero-shot learning.

\textbf{Dataset Details.} We show the video datasets used in the experiments below:

\begin{itemize}
    \item \textbf{Kinetics-400 (K400).} contains more than 230,000 10-second video clips sourced from YouTube, which have 400 categories. 
    \item \textbf{Something-Something V2 (SSv2)} covers 174 action categories. Its standard split is 168,913 training videos, 24,777 validation videos, and 27,157 testing videos. 
    \item \textbf{HMDB51} contains 7,000 videos and 51 categories. Its standard split is to train on 3570 videos and evaluate on another 1,530 videos. 
    \item \textbf{UCF-101} consists of 13,000 videos spanning 101 categories. The standard split is to train on 9,537 videos and evaluate on the left 3,783 videos.
\end{itemize}

\textbf{Implementation Details.} Tab.~\ref{tab:imple} outlines the implementation specifics for \textbf{supervised video recognition}. For \textbf{few-shot recognition}, in accordance with \cite{x-clip}, we maintain a batch size of 8 for both HMDB51 and UCF101 datasets, and set the number of training frames to 32. For \textbf{zero-shot recognition}, we initially train the model on the K400 dataset using 32 frames for 10 epochs, subsequently evaluating its performance on the test sets of HMDB51 and UCF101. All experiments are executed using 8 NVIDIA 24G 3090 GPUs.

\begin{table*}[!htb]
\caption{Comparison results of the existing competitors and OmniCLIP on \textbf{K400} dataset. The best performances are marked in \textbf{bold}. Views = \#temporal clips $\times$ \#spatial crops. The GFLOPs per view of each method is reported. The last column mentions that the trained model is suitable for zero-shot transfer.}   
    \centering
    \footnotesize
        \begin{tabular}{lcccccccc}
        \toprule
        Method & Backbone & Pre-training & Frames$\times$Views & Top-1 (\%) & Top-5 (\%)  & GFLOPs & Zero-shot \\
        \midrule
        \multicolumn{8}{l}{\textbf{Methods with Vision Training}} \\
        Uniformer-B \cite{li2023uniformer} & ViT-B/16 & IN-1k & 32$\times$4$\times$3 & 83.0 & 95.4 & 259 & \ding{55} \\
        TimeSformer \cite{TimeSformer} & ViT-B/16 & IN-21k & 96$\times$1$\times$3 & 78.0 & 93.7 & 590 & \ding{55} \\
        Mformer \cite{patrick2021keeping} & ViT-B/16 & IN-21k& 16$\times$10$\times$3 & 79.7 & 94.2 & 370 & \ding{55} \\
        Video-Swin \cite{liu2022videoswin} & Swin-B & IN-21k & 32$\times$ 4$\times$3 & 82.7 & 95.5 & 282 & \ding{55}  \\
        \midrule
        \multicolumn{8}{l}{\textbf{Methods with Vision-Language Training}} \\
        ActionCLIP \cite{wang2021actionclip} & ViT-B/16  & CLIP-400M  & 32$\times$10$\times$3 & 83.8 & 96.2 & 563 & \ding{51} \\
        X-CLIP \cite{x-clip}  & ViT-B/16 & CLIP-400M & 8$\times$4$\times$3 & 83.8 & 95.7 & 145 &\ding{51}  \\ 
        EVL \cite{evl} & ViT-B/16  & CLIP-400M & 8$\times$1$\times$3 & 82.9 & - & 444 & \ding{55} \\
        ST-Adapter \cite{st-adapter} & ViT-B/16  & CLIP-400M  & 16$\times$1$\times$3 & 82.5 & 96.0 & 911 & \ding{55} \\
        AIM \cite{aim} & ViT-B/16  & CLIP-400M  & 8$\times$1$\times$3 & 83.9 & 96.3 & 606 & \ding{55} \\
        Vita-CLIP \cite{wasim2023vita} & ViT-B/16  & CLIP-400M & 16$\times$4$\times$3 & 82.9 & 96.3 & 190 & \ding{51}  \\ 
        MotionPrompt \cite{wang2023seeing} & ViT-B/16 & CLIP-400M & 8$\times$4$\times$3 & 77.4 & 93.6 & - & \ding{51}  \\
        ILA \cite{tu2023implicit} & ViT-B/16  & CLIP-400M & 8$\times$4$\times$3 & 84.0 & 96.6 & 149 & \ding{51}  \\
        M2-CLIP \cite{wang2024m2} & ViT-B/16 & CLIP-400M & 8$\times$4$\times$3 & 83.4 & 96.3 & 214 & \ding{51}  \\
        \rowcolor{tabcolor} OmniCLIP (Ours) & ViT-B/16 & CLIP-400M & 8$\times$4$\times$3 & \textbf{84.1} & \textbf{96.7} & 130 & \ding{51}  \\
        \bottomrule        
        \end{tabular}    
    \label{tab:k400}
\end{table*}

\begin{table}
\caption
{Comparison results of the existing competitors and OminiCLIP on \textbf{SSv2} dataset. Views = \#temporal clips $\times$ \#spatial crops. The GFLOPs per view of each method is reported. The best results are marked in \textbf{bold}.}   
    \resizebox{\linewidth}{!}{
        \begin{tabular}{lcc@{\hspace{0.2cm}}c@{\hspace{0.2cm}}c@{\hspace{0.2cm}}c@{\hspace{0.2cm}}c}
        \toprule
         Method & Backbone & Pre-training & Frames$\times$Views & Top-1 (\%)  & GFLOPs \\
        \midrule
        \multicolumn{6}{l}{\textbf{Methods with Vision Training}} \\
        ViViT \cite{vivit} & ViT-L/14 & IN-21K+K400 & 8$\times$1$\times$3 & 65.4& 903 \\
        TimeSformer \cite{TimeSformer}& ViT-L/14 & IN-21K & 96$\times$1$\times$3 & 62.4  & 2380 \\
        Mformer \cite{patrick2021keeping} & ViT-B/16 & IN-21K+K400 & 16$\times$1$\times$3 & 66.5 & 370 \\        
        \midrule
        \multicolumn{6}{l}{\textbf{Methods with Vision-Language Training}} \\
        X-CLIP \cite{x-clip} & ViT-B/16  & CLIP-400M & 8$\times$4$\times$3 & 57.8  & 145\\ 
        EVL \cite{evl} & ViT-B/16  & CLIP-400M & 16$\times$1$\times$3 & 61.7  & 345 \\  
        ST-Adapter \cite{st-adapter} & ViT-B/16  & CLIP-400M & 8$\times$3$\times$1 & 67.1 & 489 \\
        AIM \cite{aim} & ViT-B/16 & CLIP-400M & 8$\times$1$\times$3 & 66.4  & 208 \\
        ILA \cite{tu2023implicit} & ViT-B/16& CLIP-400M  & 8$\times$4$\times$3 & 65.0  & 214 \\
        M2-CLIP \cite{wang2024m2} & ViT-B/16 & CLIP-400M & 8$\times$4$\times$3 & 66.9 & - \\
        \rowcolor{tabcolor} OmniCLIP (Ours) & ViT-B/16 & CLIP-400M & 8$\times$4$\times$3 & 67.0 & 128 \\
        \rowcolor{tabcolor} OmniCLIP (Ours) & ViT-B/16 & CLIP-400M & 16$\times$4$\times$3 & \textbf{67.3}  & 255 \\
        \bottomrule
        \end{tabular}    
    }
    \label{tab:ssv2}
\end{table}

\begin{table}[!tb]
    \caption{Comparison of supervised video recognition on both \textbf{HMDB51} and \textbf{UCF101} datasets. $^\dagger$ denotes the results implemented with the released codes. The best results are marked in \textbf{bold}. }  
    \resizebox{\linewidth}{!}{
        \begin{tabular}{l@{\hspace{0.1cm}}c@{\hspace{0.1cm}}c@{\hspace{0.1cm}}c@{\hspace{0.1cm}}c}
        \toprule
        \multirow{2}{*}{Method}& \multicolumn{2}{c}{HMDB-51} & \multicolumn{2}{c}{UCF-101} \\
        \cmidrule{2-3} \cmidrule{4-5}
         & \small Top-1 (\%) & \small Top-5 (\%)  & \small Top-1 (\%) & \small Top-5 (\%) \\         
        \midrule 
        I3D \cite{k400} & 74.30 & - & 95.10 & - \\
        A5 \cite{a5} & 66.40 & 92.10 & 93.60 & 99.00 \\
        Vita-CLIP \cite{wasim2023vita} $^{\dagger}$ & 71.18 & 94.12 & 93.71 & 99.50 \\
        X-CLIP \cite{x-clip} $^{\dagger}$ & 70.94 & 93.39 & 94.37 & 99.34 \\ 
        MotionPrompt \cite{wang2023seeing} & 72.90 & 93.20 & 96.30 & 99.30 \\
        \midrule
        OmniCLIP (Ours) & \textbf{76.64} & \textbf{95.89} & \textbf{96.30} & \textbf{99.56} \\
        \bottomrule
        \end{tabular}    
    }  
    \label{tab:hmdb-ucf}
\end{table}

\subsection{Results of Supervised Video Recognition.}

\textbf{Datasets and Implementation Details.}
We assess our method in a supervised video recognition setting across four benchmarks: Kinetics-400 (\textbf{K400}) \cite{k400}, Something-Something V2 (\textbf{SSv2}) \cite{goyal2017something}, HMDB51 \cite{kuehne2011hmdb}, and UCF101 \cite{soomro2012ucf101}. In all conducted experiments, the model undergoes training for a total of 50 epochs. Specifically, for the K400 and SSv2 datasets, the learning rates are set to 2e-3 and 3.5e-3, respectively. Additionally, we employ the AdamW optimizer in conjunction with a cosine annealing strategy to optimize our model. Unless otherwise stated, the input samples comprise 8 frames. For evaluation purposes, we adopt a supervised setting and utilize 4 temporal and 3 spatial views, with each view containing 8 frames.

\textbf{Results on K400.} In Tab.~\ref{tab:k400}, we conduct a thorough comparative analysis of various competitors and our OmniCLIP on the \textbf{K400} dataset, evaluating their performance in several critical aspects. Our OmniCLIP exhibits the best performance, with Top-1 and Top-5 accuracy of 84.1\% and 96.7\%, respectively. In comparison to the most related CLIP-based competitors, including ActionCLIP \cite{wang2021actionclip}, X-CLIP \cite{x-clip} Vita-CLIP \cite{wasim2023vita}, and M2-CLIP \cite{wang2024m2}, OmniCLIP demonstrates superior performance, outperforming the second-best competitor by 0.3\% and 0.4\% in Top-1 and Top-5 accuracy, respectively. Notably, these improvements are particularly noteworthy given the scale of the K400 video dataset. Moreover, OmniCLIP excels in resource efficiency, achieving a minimum of 130 GFLOPs, highlighting its ability to strike a balance between accuracy and computational cost. This makes our method an ideal candidate for deployment in resource-constrained settings. Additionally, OmniCLIP showcases its adaptability and versatility through its proficiency in zero-shot transfer learning, enabling seamless transitions to other video recognition tasks.

\begin{table}[!tb]
    \caption{Comparison results (\%) under few-shot setting. We compare OmniCLIP with approaches that explicitly adapt CLIP for video recognition on \textbf{HMDB51} and \textbf{UCF101}. The best results are marked in \textbf{bold}.}
    \resizebox{\linewidth}{!}{    
        \begin{tabular}{lcccc|cccc}
        \toprule
        \multirow{2}{*}{Method} & \multicolumn{4}{c}{HMDB51} & \multicolumn{4}{c}{UCF101} \\
        \cmidrule{2-9}
             & K=2 & K=4 & K=8 & K=16 & K=2 & K=4 & K=8 & K=16 \\
        \midrule
            Vanilla CLIP \cite{clip} & 41.9 & 41.9 & 41.9 & 41.9 & 63.6 & 63.6 & 63.6 & 63.6\\
            A5 \cite{a5} & 39.7 & 50.7 & 56.0 & 62.4 & 71.4 & 79.9 & 85.7 & 89.9\\    
            ActionCLIP \cite{wang2021actionclip} & 47.5 & 57.9 & 57.3 & 59.1 & 70.6 & 71.5 & 73.0 & 91.4\\
            XCLIP \cite{x-clip} & 53.0 & 57.3 & 62.8 & 64.0 & 76.4 & 83.4 & 88.3 & 91.4 \\
            MotionPrompt \cite{wang2023seeing}   & \textbf{55.3} & \textbf{58.7} & 64.0 & 64.6 & \textbf{82.4} &  \textbf{85.8} & 89.1 & 91.6 \\
            \rowcolor{tabcolor} OmniCLIP & 54.1 & 58.3 & \textbf{67.3} & \textbf{74.4} & 76.7 & 84.5 & \textbf{91.5} & \textbf{95.1} \\
        \bottomrule
        \end{tabular}
        }
    \label{tab:fsl-results}
\end{table}

\textbf{Results on SSv2.} SSv2 has richer temporal information and more details of action descriptions. We select recent 9 competitors for comparison. The comparison results are shown in Tab.~\ref{tab:ssv2}. We train the video encoder with a fixed classifier following previous works \cite{tu2023implicit}. Our OmniCLIP has demonstrated exceptional performance on the SSv2 dataset, achieving a Top-1 accuracy of 67.0\% while maintaining a low computational cost of 128 GFLOPs. This impressive efficiency underscores OmniCLIP's ability to effectively capture dynamic video representations. When compared to the ST-Adapter \cite{st-adapter}, which achieves a slightly higher accuracy of 67.1\% but with a significantly greater computational cost of 489 GFLOPs, OmniCLIP clearly demonstrates its superiority in balancing accuracy and computational efficiency. Furthermore, OmniCLIP exhibits remarkable adaptability, as its Top-1 accuracy improves to 67.3\% when processing 16 frames on the SSv2 dataset. While this enhancement incurs a higher computational cost of 255 GFLOPs, it remains significantly lower than the ST-Adapter, highlighting OmniCLIP's flexibility under varying resource constraints. In summary, OmniCLIP offers an outstanding balance between high accuracy and low computational costs on the SSv2 dataset, making it an ideal choice for efficient video recognition tasks.

\textbf{Results on HMDB51 and UCF101.} Tab.~\ref{tab:hmdb-ucf} presents the comparison results of five competitors and our OmniCLIP on both the HMDB51 and UCF101 datasets. Notably, all methods except for I3D, are based on the ViT-B/16 architecture. Our OmniCLIP consistently outperforms existing competitors across both Top-1 and Top-5 metrics on both datasets. On the HMDB51 dataset, OmniCLIP achieves a remarkable Top-1 accuracy of 76.64\% and a Top-5 accuracy of 95.89\%. This performance surpasses the second-best method, MotionPrompt \cite{wang2023seeing}, by significant margins of 3.74\% and 2.69\% in Top-1 and Top-5 accuracy, respectively. This underscores the superiority of our approach in capturing and representing the subtleties of human actions in this challenging dataset. Similarly, on the UCF101 dataset, OmniCLIP matches the highest reported Top-1 accuracy of 96.30\% and sets a new benchmark with a Top-5 accuracy of 99.56\%.

\begin{table}[!tb]
    \caption{Comparison (\%) for zero-shot performances on both \textbf{HMDB51} and \textbf{UCF101} datasets. The best results are marked in \textbf{bold}.}      
    \centering
        \begin{tabular}{l@{\hspace{1cm}}|@{\hspace{0.5cm}}c@{\hspace{0.5cm}}|@{\hspace{0.5cm}}c}
        \toprule
        Method & HMDB51 & UCF101 \\
        \midrule
        \multicolumn{3}{l}{\textbf{Methods with Vision Training}} \\   
        TS-GCN \cite{TS-GCN} & 23.2 ± 3.0 & 34.2 ± 3.1 \\
        E2E \cite{e2e} & 32.7 ± 0.0 & 48.0 ± 0.0 \\
        ER-ZSRA \cite{ER-ZSAR} & 35.3 ± 4.6 & 51.8 ± 2.9 \\
        \midrule
        \multicolumn{3}{l}{\textbf{Methods with Vision-Language Training}} \\
        Vanilla CLIP \cite{clip} & 40.8 ± 0.3 & 63.2 ± 0.2 \\
        ActionCLIP \cite{wang2021actionclip} & 40.8 ± 5.4 & 58.3 ± 3.4 \\ 
        A5 \cite{a5} &  44.3 ± 2.2 & 69.3 ± 4.2 \\
        X-CLIP B/16 \cite{x-clip} & 44.6 ± 5.2 & 72.0 ± 2.3\\
        Vita-CLIP B/16 \cite{wasim2023vita} &  48.6 ± 0.6 & 75.0 ± 0.6 \\
        MotionPrompt \cite{wang2023seeing} & 50.1 ± 5.4 & \textbf{76.4 ± 2.5} \\
        \rowcolor{tabcolor} OmniCLIP & \textbf{51.3 ± 1.2} & 73.2 ± 1.0 \\
        \bottomrule
        \end{tabular}    
    \label{tab:zs-exp}
\end{table}

\begin{figure*}[!htb]
    \centering
    \setlength{\tabcolsep}{1.5pt}
    \begin{tabular}{cccc@{\hskip 10pt}cccc@{\hskip 10pt}cccc}
        \multicolumn{12}{c}{\begin{overpic}[width=\linewidth]{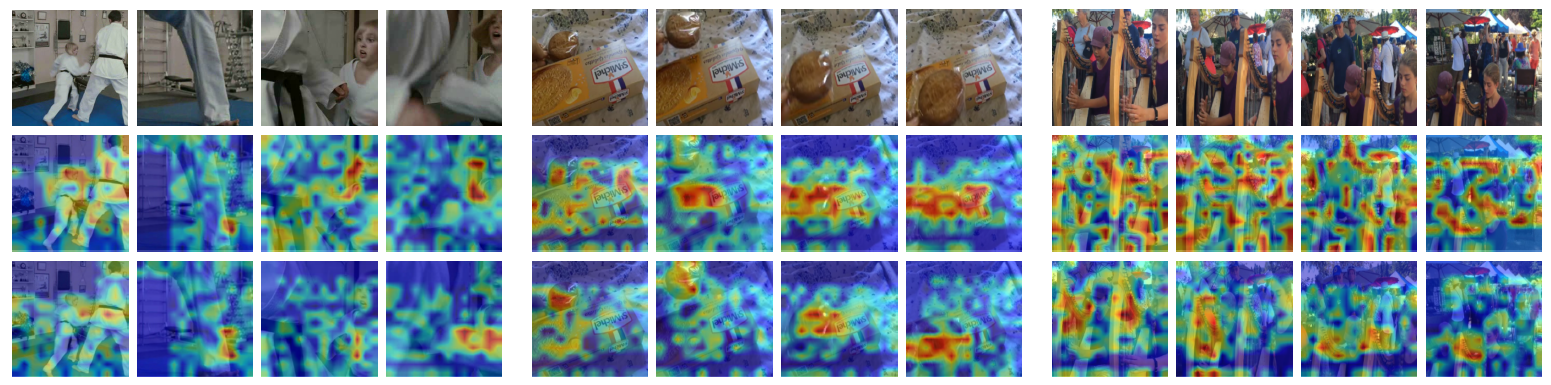}\end{overpic}} \\
    \end{tabular}
    \put(-460, -68){\small (a) Class: ``punch”}
    \put(-310, -68){\small (b) Class: ``pushing something...”}
    \put(-130, -68){\small (c) Class: ``playing harp”}
    \caption{The attention map on sample videos, showing raw frames (the first row), heatmap with vanilla CLIP (the second row), and with our OmniCLIP (the last row). The actions like 'punch', 'pushing something from behind of something', and 'playing harp' are shown.}
    \label{fig:heatmap}
\end{figure*}

\subsection{Results of Few-Shot Classification.}
\textbf{Datasets and Implementation Details.} Following \cite{x-clip}, we conduct few-shot experiments on HMDB51 and UCF101 datasets. For each category, we randomly sample K instances for model training and use the test set for evaluation. 

\textbf{Performance.} Tab.~\ref{tab:fsl-results} showcases the comparison results of our OmniCLIP against five competitors on both the HMDB51 and UCF101 datasets. Notably, across various settings of K (the number of labeled samples per class), OmniCLIP consistently demonstrates competitive performance, particularly achieving the second-best results when K is set to 2 and 4. More importantly, as K increases, OmniCLIP's advantage becomes increasingly evident. Specifically, at K~=~8 and K~=~16, OmniCLIP achieves remarkable Top-1 accuracy rates of 67.3\% and 74.4\% on the HMDB51 dataset, surpassing the runner-up by significant margins of 3.3\% and 9.8\%, respectively. On the UCF101 dataset, OmniCLIP maintains its superiority, achieving outstanding accuracy rates of 91.5\% and 95.1\% at K~=~8 and K~=~16, respectively, outperforming the second-best method by 2.4\% and 3.5\%. It is noteworthy that, at K~=~16, OmniCLIP even outperforms models that are trained with the full training data on the HMDB51 dataset (as shown in Tab.~\ref{tab:hmdb-ucf}). This underscores the robustness and efficiency of our OmniCLIP approach in effectively capturing valuable temporal and spatial information, even with limited labeled data.

\subsection{Results of Zero-Shot Classification.}
\textbf{Datasets and Implementation Details.} Following \cite{x-clip}, we firstly train our model on K400, and evaluate the zero-shot transfer capability on two datasets: HMDB51 and UCF101. We follow \cite{zoph2018learning} and report average top-1 accuracy and standard deviation on three splits of the test set.

\begin{table*}[htb]

\begin{minipage}[t]{0.32\textwidth}
\makeatletter\def\@captype{table}
\caption{Impact (\%) of different modules.}
\begin{tabular}{cc|c@{\hspace{0.1cm}}c}
    \toprule
        \small \textbf{PTA} & \small \textbf{SPG} & \small {K400-tiny} & \small {HMDB51} \\
    \midrule
        \ding{55} &\ding{55}  & 51.60 & 40.90 \\
         \ding{51} & \ding{55}  & 76.40 & 75.23\\
         \ding{55} & \ding{51} & 73.20 & 71.34 \\
         \ding{51} & \ding{51} & 77.20 & 76.64\\
    \bottomrule
\end{tabular}

\label{tab:ab-modules}
\end{minipage}
\begin{minipage}[t]{0.32\textwidth}
\caption{Impact (\%) of temporal ratio.}
\makeatletter\def\@captype{table}
\centering
\begin{tabular}{c|c@{\hspace{0.1cm}}c}
    \toprule
        \small \textbf{Ratio} & \small {K400-tiny} & \small {HMDB51} \\
    \midrule
        1/8 & 76.84 & 75.34\\
        1/4 & 77.20 & 76.64 \\
        1/2 & 77.13 & 76.34\\
        1   & 76.93 & 75.94\\
    \bottomrule
\end{tabular}

\label{tab:ab-ratio}
\end{minipage}   
\begin{minipage}[t]{0.32\textwidth}
\caption{Impact (\%) of self-prompt.}
\makeatletter\def\@captype{table}
\begin{tabular}{c|c@{\hspace{0.1cm}}c}
    \toprule
        \small \textbf{Model} & \small {K400-tiny} & \small {HMDB51} \\
    \midrule
        Avg. & 76.20 & 75.64 \\
        Max. & 76.03 & 75.10\\
        Max. + Projector & 76.94 & 76.23 \\
        Avg. + Projector & 77.20 & 76.64 \\
    \bottomrule
\end{tabular}

\label{tab:ab-spatial-prompts}
\end{minipage}   
\end{table*}

\textbf{Performance.} Tab.~\ref{tab:zs-exp} presents the results of our zero-shot experiments. From the results, OmniCLIP emerges as a standout performer. On the UCF101 dataset, OmniCLIP achieves an outstanding average Top-1 accuracy of 73.2\%, surpassing the vanilla CLIP model by a significant margin of 10.0\%. Moreover, OmniCLIP excels on the HMDB51 dataset, achieving the best performance with a Top-1 accuracy of 51.3\%.  This further underscores the generalizability of our method across diverse video recognition tasks. However, it's worth noting that OmniCLIP's performance on UCF101 is slightly inferior to some other methods. We attribute this to the dataset's strong biases towards appearance and objects, which may limit the effectiveness of our approach in capturing temporal information and dependencies. Nevertheless, on datasets where temporal cues are more critical, such as HMDB51, OmniCLIP excels and demonstrates its superiority.

\subsection{Further Analysis}
To assess the impact of each component within OmniCLIP, we conduct ablation studies on the K400-tiny (a smaller split of the full K400) and HMDB51 datasets. Additionally, we provide visualizations for further insights.

\textbf{Impacts of different modules.} Tab.~\ref{tab:ab-modules} presents a comparison between the effects of the Parallel Temporal Adapter (PTA) and the Self-Prompt Generator (SPG). Incorporating the PTA alone leads to a notable increase in performance, achieving 76.40\% on K400-tiny and 75.23\% on HMDB51. Similarly, the implementation of SPG independently results in enhanced outcomes, recording 73.20\% on K400-tiny and 71.34\% on HMDB51. The most significant improvement is observed when PTA and SPG are used in conjunction, culminating in 77.20\% on K400-tiny and 76.64\% on HMDB51. This indicates the synergistic benefit of integrating temporal, spatial, and dynamic spatial-temporal omni-scale features.

\begin{table}[!t]
 \caption{Impact (\%) of different temporal adapter locations.}
    \resizebox{0.9\linewidth}{!}{
        \begin{tabular}{cccc@{\hspace{0.2cm}}|c@{\hspace{0.3cm}}c}
        \toprule
            \multicolumn{4}{c|}{\textbf{PTA layers}} & \multirow{2}{*}{{K400-tiny}} &  \multirow{2}{*}{{HMDB51}} \\
            1-3 & 4-6 & 7-9 & 10-12 &  &\\
        \midrule
            \ding{55} & \ding{55} & \ding{55} & \ding{55} & 51.60 & 40.90 \\
            \ding{51} &  \ding{55} & \ding{55}  & \ding{55}  & 75.20 & 74.35 \\
            \ding{51} & \ding{51} &  \ding{55}&  \ding{55}& 76.41 & 75.83 \\
            \ding{51} & \ding{51} & \ding{51} &  \ding{55}& 77.01 & 76.34 \\
            \ding{51} & \ding{51} & \ding{51} & \ding{51} & 77.20 & 76.64\\
        \bottomrule
        \end{tabular}
    }   
    \label{tab:ab-locations}
\end{table}

\textbf{Impacts of temporal ratio.} The temporal ratio refers to the compression ratio used by the Down projector in Eq.~\ref{eq:temporal prompts}, and it plays a crucial role in determining the efficiency of video recognition. Tab.~\ref{tab:ab-ratio} provides a comprehensive analysis of how varying these ratios within the temporal adapter affects recognition performance. Our findings indicate that a ratio of 1/4 yields optimal results, achieving the highest scores of 77.20\% on K400-tiny and 76.64\% on HMDB51. When the ratio deviates from this optimal value, either towards a lower value (1/8) or a higher one (1/2, 1), we observe a decrease in performance. This suggests that maintaining a balanced temporal ratio is crucial for achieving the best recognition outcomes. We hypothesize that overly complex temporal modules may lead to overfitting, which in turn degrades overall performance.

\textbf{Impacts of self-prompt.}  
Tab.~\ref{tab:ab-spatial-prompts} illustrates the impact of various prompt-generation methods on video recognition performance. The \textbf{Avg.} method utilizes average pooling, and the \textbf{Max.} approach applies max pooling for prompt generation. Additionally, incorporating a \textbf{Projector} represents an advanced refinement of these prompts via a learnable mechanism. The findings indicate that both average and max pooling independently deliver robust results, achieving Top-1 accuracies of 75.65\% and 75.10\%, respectively. This reflects their effectiveness in capturing key video content features. Moreover, the integration of the learnable projector with both \textbf{Avg.} and \textbf{Max.} further enhance their performance. We hypothesize that the \textbf{Avg.} is better suited for capturing smoother and more representative spatial features, which are crucial for video recognition, in contrast to the \textbf{Max.} Overall, the self-prompts can effectively integrate with frozen visual features, thereby adeptly capturing spatial cues.

\textbf{Impacts of different temporal adapter locations.} We explore the effects of different placements of the parallel temporal adapter within the visual branch of the ViT-B/16 model, which is divided into 12 blocks. These blocks are grouped into four sets, each containing three blocks. By strategically placing the temporal adapter in various groups, we assess the effect of how its location influences the model's performance, as detailed in Tab.~\ref{tab:ab-locations}. Inserting the adapter into the first group (blocks 1-3) results in a substantial improvement, with performance scores reaching 75.20\% on K400-tiny and 74.35\% on HMDB51. Notably, when the temporal adapter is extended across all four groups, OmniCLIP obtains its peak performance, demonstrating enhancements of 25.60\% and 35.74\% over the vanilla CLIP model on K400-tiny and HMDB51, respectively. This underlines the significant impact of the temporal adapter's distribution in the visual branch, showcasing a progressive enhancement in video recognition capabilities for capturing the temporal cues enhancement.

\textbf{Visualization.} Fig.~\ref{fig:heatmap} illustrates the visual attention of the OmniCLIP. Videos are samples from various datasets for comparison and the attention map of the [CLS] token from the last layer is presented. The results demonstrate that OmniCLIP tends to focus on the moving objects (\textit{e.g.} bread, hands) and the more important object (\textit{e.g.} harp) of recognition, while vanilla CLIP is confused by the multi-objects and background across frames.

\begin{figure}[!htb]
    \centering
   \begin{overpic}[width=\linewidth,height=0.4\linewidth]{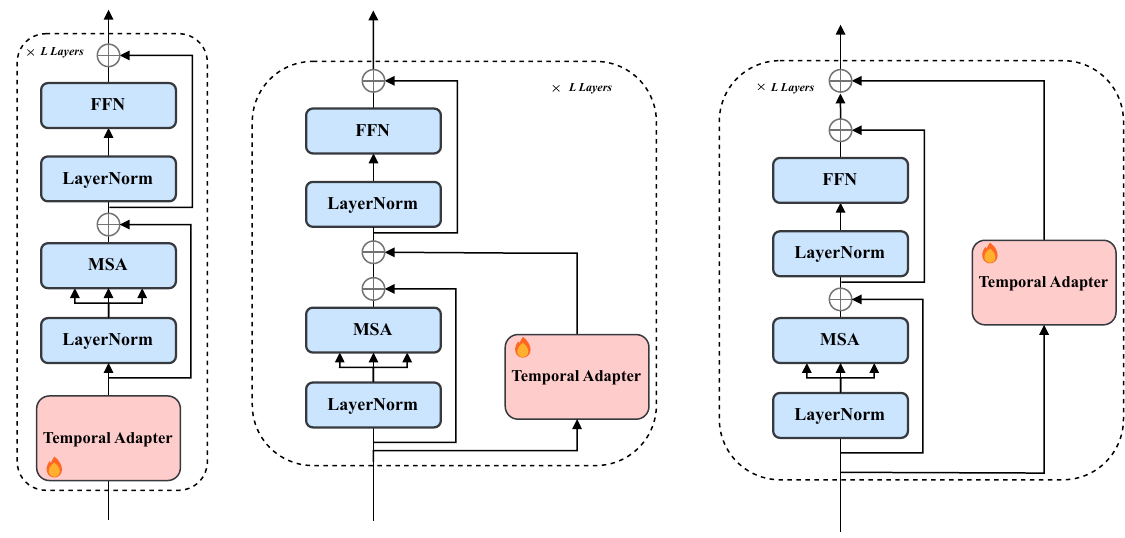} \small
   \put(2,-5){(a)~Cascade.}
   \put(28,-5){(b)~Attention Parallel.}
   \put(65,-5){(c)~Block Parallel~(Ours).}
   \end{overpic}   
   \vspace{5mm}
    \caption{The different combinations of the temporal adapter.}
    \label{fig:ap-framwork1}
\end{figure}

\begin{table}[!htb]
    \centering
     \caption{Impact (\%) of the combinations of the temporal adapter.}
    \begin{tabular}{c@{\hspace{0.5cm}}|c@{\hspace{0.6cm}}c}
        \toprule
            \small \textbf{Model} & \small {K400-tiny} & \small {HMDB51} \\
        \midrule
            Cascade & 75.35 & 75.23 \\
            Attention Parallel & 76.51 & 75.51\\
            Ours & 77.20 & 76.64 \\
        \bottomrule
    \end{tabular}
    \label{tab:combinition-ta}
\end{table}

\textbf{Different Combinations of Temporal Blocks.} In this experiment, we assess various methods for integrating temporal adapters, including the cascade connection (Fig.~\ref{fig:ap-framwork1}~(a)), the attention parallel connection (Fig.~\ref{fig:ap-framwork1}~(b)), and our proposed parallel temporal adapter (Fig.~\ref{fig:ap-framwork1}~(c)). Notably, the cascade variant of the temporal adapter can be viewed as a specialized form of the ST-Adapter, leveraging 3D convolution to capture spatial-temporal information. The results of these methodologies are summarized in Tab.~\ref{tab:combinition-ta}. While both the cascade and attention parallel frameworks demonstrate their abilities in temporal and spatial modeling, resulting in performance improvements on the HMDB51 and UCF101 datasets, our proposed parallel temporal adapter still surpasses these strategies. Given the cascade architecture's relatively higher computational cost, we can confidently conclude that our proposed parallel temporal adapter not only outperforms other integration methods in terms of performance but also offers greater efficiency.

\section{Conclusion}
In this paper, we have presented OmniCLIP, an innovative approach that adapts CLIP model for video recognition by incorporating spatial, temporal, and dynamic spatial-temporal omni-scale features. We have designed a parallel temporal adapter (PTA) to establish effective temporal modeling, thus filling a crucial gap in the vanilla CLIP model specifically for video processing. Additionally, we have developed a self-prompt generator module (SPG) that refines spatial scale features in videos. The synergy of PTA and SPG enables OmniCLIP to effectively capture dynamic spatial-temporal features. Experimental evaluations across a range of benchmarks consistently have demonstrated that OmniCLIP excels in learning omni-scale video features, leading to notable improvements, especially in few-shot video recognition scenarios.

\section*{Acknowledgment}
This research was supported in part by Zhejiang Provincial Natural Science Foundation of China under Grant LD24F020016, the Key R\&D  Program of Zhejiang Province, China 2023C01043, NSFC (62002320, U19B2043, 52305590), Science and Technology Innovation 2025 Major Project of Ningbo (2023Z236).

\bibliography{ref}
\end{document}